%% file: main.tex
\documentclass{article}

\usepackage{microtype}
\usepackage{graphicx}
\usepackage{subfigure}
\usepackage{booktabs} % for professional tables

\usepackage{hyperref}
\usepackage{pgfplots}

\usepackage[accepted]{icml2024}

\usepackage{amsmath}
\usepackage{amssymb}
\usepackage{mathtools}
\usepackage{amsthm}
\usepackage{natbib}
\usepackage{graphicx}
\usepackage{multirow}
\usepackage{tabularx}

\usepackage[capitalize,noabbrev]{cleveref}

\theoremstyle{plain}

\theoremstyle{definition}

\theoremstyle{remark}

\usepackage[textsize=tiny]{todonotes}

\icmltitlerunning{Representing Online Handwriting for Recognition}

\begin{document}

\twocolumn[
\icmltitle{Representing Online Handwriting for Recognition \\
in Large Vision-Language Models}

% It is OKAY to include author information, even for blind
% submissions: the style file will automatically remove it for you
% unless you've provided the [accepted] option to the icml2024
% package.

% List of affiliations: The first argument should be a (short)
% identifier you will use later to specify author affiliations
% Academic affiliations should list Department, University, City, Region, Country
% Industry affiliations should list Company, City, Region, Country

% You can specify symbols, otherwise they are numbered in order.
% Ideally, you should not use this facility. Affiliations will be numbered
% in order of appearance and this is the preferred way.
\icmlsetsymbol{equal}{*}
\icmlsetsymbol{tech_lead}{§}

\begin{icmlauthorlist}
\icmlauthor{Anastasiia Fadeeva}{equal,gr}
\icmlauthor{Philippe Schlattner}{equal,gr}
\icmlauthor{Andrii Maksai}{tech_lead,gr}
\icmlauthor{Mark Collier}{gr}
\icmlauthor{Efi Kokiopoulou}{gr}
\icmlauthor{Jesse Berent}{gr}
\icmlauthor{Claudiu Musat}{tech_lead,gr}
%\icmlauthor{}{sch}
%\icmlauthor{}{sch}
\end{icmlauthorlist}

\icmlaffiliation{gr}{Google Research}
%\icmlaffiliation{comp}{Company Name, Location, Country}
%\icmlaffiliation{sch}{School of ZZZ, Institute of WWW, Location, Country}

\icmlcorrespondingauthor{{Anastasiia Fadeeva}}{fadeich@google.com}
\icmlcorrespondingauthor{{Andrii Maksai}}{amaksai@google.com}

% You may provide any keywords that you
% find helpful for describing your paper; these are used to populate
% the "keywords" metadata in the PDF but will not be shown in the document
%\icmlkeywords{Machine Learning, ICML}
\icmlkeywords{Handwriting recognition, Machine Learning}

\vskip 0.3in
]

% this must go after the closing bracket ] following \twocolumn[ ...

% This command actually creates the footnote in the first column
% listing the affiliations and the copyright notice.
% The command takes one argument, which is text to display at the start of the footnote.
% The \icmlEqualContribution command is standard text for equal contribution.
% Remove it (just {}) if you do not need this facility.

%\printAffiliationsAndNotice{}  % leave blank if no need to mention equal contribution
\printAffiliationsAndNotice{\icmlEqualContribution} % otherwise use the standard text.

\begin{abstract}
The adoption of tablets with touchscreens and styluses is increasing, and a key feature is  converting handwriting to text, enabling search, indexing, and AI assistance.

Meanwhile, vision-language models (VLMs) are now the go-to solution for image understanding, thanks to both their state-of-the-art performance across a variety of tasks and the simplicity of a unified approach to training, fine-tuning, and inference. While VLMs obtain high performance on image-based tasks, they perform poorly on handwriting recognition when applied naively, i.e., by rendering handwriting as an image and performing optical character recognition (OCR). 

In this paper, we study online handwriting recognition with VLMs, going beyond naive OCR. We propose a novel tokenized representation of \textbf{digital ink} (online handwriting) that includes both a time-ordered sequence of strokes as text, and as image. We show that this representation yields results comparable to or better than state-of-the-art online handwriting recognizers. Wide applicability is shown through results with two different VLM families, on multiple public datasets. Our approach can be applied to off-the-shelf VLMs, does not require any changes in their architecture, and can be used in both fine-tuning and parameter-efficient tuning. We perform a detailed ablation study to identify the key elements of the proposed representation.

\end{abstract}

\input{intro}

\input{background}

\input{method}
\input{results}
\input{ablation}
\input{related_work}
\input{conclusion}

\newpage
\bibliography{cite}
\bibliographystyle{icml2024}

%%%%%%%%%%%%%%%%%%%%%%%%%%%%%%%%%%%%%%%%%%%%%%%%%%%%%%%%%%%%%%%%%%%%%%%%%%%%%%%
%%%%%%%%%%%%%%%%%%%%%%%%%%%%%%%%%%%%%%%%%%%%%%%%%%%%%%%%%%%%%%%%%%%%%%%%%%%%%%%
% APPENDIX
%%%%%%%%%%%%%%%%%%%%%%%%%%%%%%%%%%%%%%%%%%%%%%%%%%%%%%%%%%%%%%%%%%%%%%%%%%%%%%%
%%%%%%%%%%%%%%%%%%%%%%%%%%%%%%%%%%%%%%%%%%%%%%%%%%%%%%%%%%%%%%%%%%%%%%%%%%%%%%%
\newpage
\appendix
\onecolumn
\input{appendix}

\end{document}

%% file: intro.tex
\section{Introduction}

Digital alternatives to writing on paper are expanding. One of the key features users need is to seamlessly transition between modalities and turn their handwritten notes into printed text \cite{WeMayInk}. This feature depends on the quality of the underlying handwriting recognition models. 

\textbf{Historical perspective.} Approaches to handwriting recognition have evolved over time alongside similar problems in speech recognition and OCR, going from segment-and-decode models \cite{541414} to RNNs \cite{carbune2020recognition, Graves2009ANC} to Transformer-based approaches \cite{MSdocTr}. However, as in other modalities, it is still far from being solved, especially in more complex cases that involve whole-page note-taking, math expression recognition, and scripts with small amounts of training data. This is also visible through the variety of model architectures used to solve these problems~\cite{ICDAR_CROHME}.

\textbf{Why use VLMs?} LLMs and VLMs \cite{brown2020language, touvron2023llama, chen2023palix} have shown impressive performance in a variety of tasks and across different modalities and they can offer multiple benefits if they can be used for solving handwriting recognition problems (or any other target domain). The obvious one is a potential quality improvement coming from their scale and underlying language model. Their simple design allows fine-tuning a single model end-to-end  using standard and widely available tools in contrast to standard multi-step recognition models, ex.~\cite{carbune2020recognition}.  Finally, they allow seamlessly mixing multiple handwriting tasks expanding the plethora of tasks they can already perform.

\textbf{How to use VLMs?} To be able to instill VLMs with handwriting recognition, one needs a representation of digital ink that is suitable for use with VLMs. A straightforward approach is to simply provide a rendering of digital ink as an input image and perform OCR. However, for handwriting recognition, OCR-only performance is subpar to the quality of specialized online handwriting recognition models that operate on the digital ink as a time-ordered sequence of points~\cite{ICDAR_CROHME}.

\textbf{Our work}. The focus of our work is the representation of digital ink for VLMs that is applicable across different datasets and model families, and yields comparable performance to state-of-the-art task-specific models. To our knowledge, we are the first to explore stroke-based representations in VLMs for handwriting recognition. 

In our search for a widely applicable representation of  digital ink we explore two main options: based on images and time-ordered sequences of points. We look for the optimal way of rendering ink into an image and of discretizing the sequence of points into a sequence of text tokens that can be consumed by VLMs. We show how these representations should be combined together to achieve optimal performance. 

We find that it is possible to obtain good recognition quality while representing digital ink as text. This is unlike some other modalities like audio~\cite{rubenstein2023audiopalm}, where adding a new modality into an existing model requires extending the token dictionary with the tokens of the new modality as well as modifying the model architecture. This means that our approach does not require any changes to existing models and enables adding handwriting recognition capabilities to pre-trained VLMs by fine-tuning or even parameter-efficient tuning, which further preserves original capabilities of the model. Our findings generalize across two model families and several different handwriting recognition datasets.

To sum up, our main contributions are as follows:
\begin{itemize}
\item We propose a representation for digital ink that combines images and time-ordered token sequences, which is suitable for use with VLMs;
\item We show that such a dual representation is vital for achieving performance comparable to state-of-the-art task-specific handwriting recognition models; to our knowledge this is the first work to explore stroke-based representations for online handwriting recognition in VLMs;
\item We show that our representation is suitable for fine-tuning or parameter-efficient tuning of a pre-trained VLM and does not require changing model architecture or vocabulary;
\item We perform a detailed ablation study to find the best way of representing digital ink as images and as token sequences.
\end{itemize}

%% file: background.tex
\section{Background}
\label{sec:backdround}

This paper focuses on \textbf{online handwriting recognition}, meaning the input includes spatial and time information. Related work on the topic is summarized in Section~\ref{sec:related_work}. We denote a \textbf{stroke} $s$ as a sequence of triplets $(x, y, t)$ where $x$ and $y$ are coordinates on the screen and $t$ is time information \cite{carbune2020recognition}.
We denote an \textbf{ink} $\text{I} = [s_0, \ldots, s_n]$ as a sequence of written strokes (aka pen-down strokes). The \textbf{input} of our model is an ink $\text{I}$ and the \textbf{output} is the text written in the ink.

\begin{figure}[!ht]
    \centering
    \includegraphics[width=\textwidth/2]{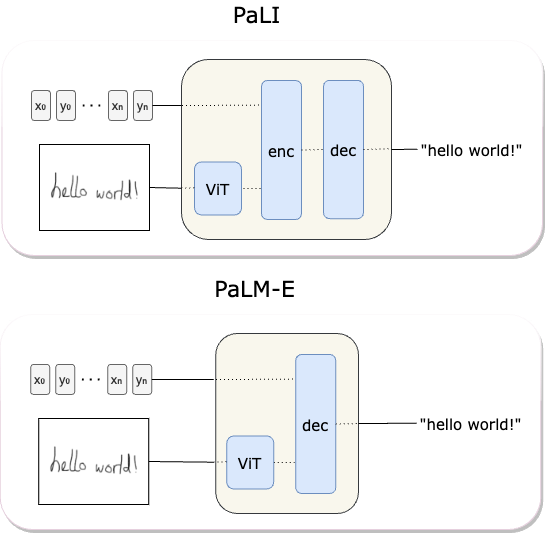}
     \vspace{-20pt}
    \caption{PaLI and PaLM-E architectures for ink recognition.}
    \label{fig:architectures}
\end{figure}

The \textbf{VLM} architectures that we use in this work are PaLI \cite{chen2023pali, chen2023palix} and PaLM-E \cite{driess2023palme}, transformer-based models~\cite{DBLP:journals/corr/VaswaniSPUJGKP17}. PaLI is an encoder-decoder model that combines image and text representations in the encoder as shown in Fig.~\ref{fig:architectures}. PaLM-E is a decoder-only model that combines image and textual embeddings in its input. The main difference between the two is the presence of transformer encoder in PaLI. In our experiments with PaLM-E, we use non-causal mask on the the input as described in \citet{pretrainingLM} given that \citet{10.5555/3455716.3455856} has shown that this approach provides similar quality to encoder-decoder models. 

\textbf{Text tokenization in VLMs} plays an important role in the success of the training. Most models use the Byte-Pair Encoding (BPE) procedure \cite{Kudo2018SentencePieceAS}, however there are different approaches to tokenizing numbers in text. It was shown in \citet{liu2023goat} that tokenizing each digit separately helps with arithmetic operations, at the price of making the input sequence longer, as opposed to a BPE vocabulary that contains multi-digit tokens. We consider both options as PaLI utilizes standard BPE and PaLM-E uses the single digit tokenization strategy.

%% file: method.tex
\section{Method}
\label{sec:representation}

In our work we focus on representation choices that make online handwriting recognition suitable for VLMs. In this section we describe a range of possibilities to represent online handwriting as a sequence of time-ordered tokens and as images. We provide the recipe that we used used in the experiments and the intuition behind it. 

\begin{figure*}[!ht]
    \centering
    \includegraphics[width=\textwidth]{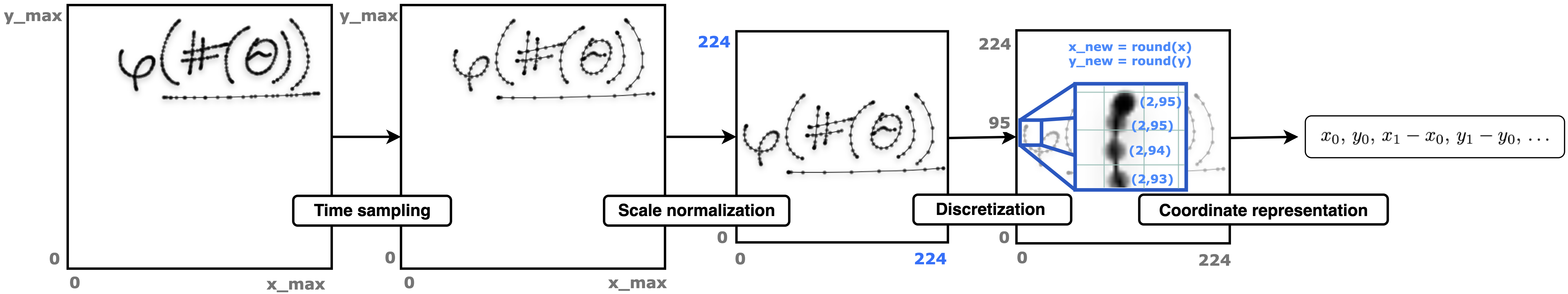}
    \vspace{-20pt}
    \caption{The full pipeline for the \textbf{sequence representation} in VLMs. This pipeline includes time sampling, scale normalization, discretization with uniform grid and representation of points with two coordinates in text.}
    \label{fig:coordinate_tokenization}
\end{figure*}
\subsection{Sequence representation}
\label{sec:sequence_representation}

In its original form online inks are represented as sequences of strokes with coordinates $x, y$ and time information $t$, sampled. This could be encoded in text format as: ``$\text{stroke}~~x_{0, 0}~~y_{0, 0}~~t_{0, 0}~~x_{0, 1}~~y_{0, 1}~~t_{0, 1} \ldots$''. Even though this representation can be directly fed into a VLM, we found it crucial to consider time sampling, scale normalization, coordinate representation, discretization, and token dictionary. Below we outline each of these and Figure~\ref{fig:coordinate_tokenization} depicts them.

Time sampling, scale normalization, and token dictionary choice have an immediate effect on the length of the sequence used to represent this ink and Table~\ref{table:time_norm} shows how doing them allows to reduce the sequence length significantly. The effect of other choices, namely coordinate representation, codebook, and token dictionary, is studied in Section~\ref{sec:ablation_tokenization}.

\textbf{Time sampling.} In order to normalize sampling frequency among different devices and reduce the sequence length, we resample points at regular time intervals within each stroke. It is critical to pick an appropriate time delta because with high values an ink can lose important writing details (see Appendix~\ref{sec:appendix_time_sampling} for examples). After resampling, all points within each stroke have the same time delta between each other, so we can omit value $t$ for the text representation (information about the duration of time between strokes is thus discarded). This allows us to significantly 
shorten the sequences due to lower number of points and an ability to omit value $t$ from text, as seen in Tab.~\ref{table:time_norm}.

\begin{table}[!ht]
\caption{Effect of normalization and tokenization on the sequence length. Median numbers with mT5~\cite{xue2021mt5} tokenizer on MathWriting dataset \cite{MathWriting}. The first line reports the length when representing the ink as the original sequence of $x$, $y$, and $t$ coordinates as described in the beginning of Section~\ref{sec:sequence_representation} and each following line cumulatively adds some form of processing described in the section. After time sampling, we remove the $t$ from representation as described in the appropriate paragraph, and after scale normalization, we round $x$ and $y$ to the nearest integer.}
\label{table:time_norm}
%\small
\centering
\begin{tabular}{|l|c|c|} 
 \hline
 Representation & \# Points & \# Tokens \\
 \hline
 Original x,y,t & 313 & 2692\\
 \hline
 +Time sampling & 178 & 1954 \\
 \hline
 +Scale normalization & 178 & 381 \\
 \hline
 +Extended token dictionary & 178 & 367 \\
 \hline
\end{tabular}
\end{table}

\textbf{Scale normalization.} We scale and shift, preserving the aspect ratio, so that all points fit into the range between 0 and $N$ (where $N$ is the ViT encoder image size). That helps to account for potential differences in the input canvas size and to reduce the sequence length through smaller point coordinates that have a more compact text representation. We use the information about the scaling range when doing discretization, as described below.

\textbf{Coordinate representation.} We represent points of the ink using the offsets of coordinates from one time-step to the next rather than the absolute coordinates:
$(x_t^r, y_t^r) = (x_t, y_t) - (x_{t-1}, y_{t-1})$.
In the ablation study, we compare the absolute and relative representations. See Fig.~\ref{fig:coordinate_tokenization} for more details.

\textbf{Discretization codebook.} We use two values to represent each point $(x, y)$ in the ink, by rounding the normalized $x$ and $y$ coordinate to the nearest integer. In the ablation study, we show that this approach leads to better accuracy than using a learned codebook of offsets, similar to \citet{ribeiro2020sketchformer}, where each offset is represented by a single token.

\textbf{Token dictionary.} We represent each point $(x, y)$ directly as text, as opposed to extending the token dictionary \cite{rubenstein2023audiopalm}. We use a separator expression to indicate the beginning of the new stroke, as ``$<$stroke$>$ 2 1 2 2 \ldots $<$stroke$>$ 3 1 3 2 3 3 \ldots''. In the ablation study, we compare this to the approach where the token dictionary of the model is extended with new tokens, which has similar performance but requires changes to the model.

\begin{figure*}[!ht]
    \centering
    \includegraphics[width=\linewidth]{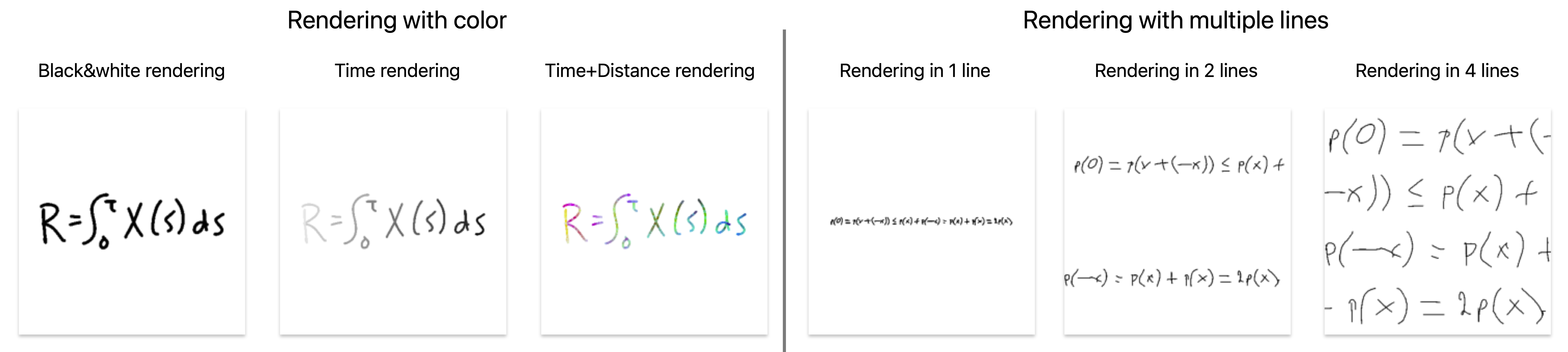}
    \vspace{-20pt}
    \caption{Examples of different rendering options. Rendering options for color – black\&white, time from Eq.~\ref{eq:speed_rendering} and time+distance from Eq.~\ref{eq:speed_rendering}. Examples of rendering in one, two or four lines.}
    \label{fig:rendering}
\end{figure*}

Note that when the input is provided as text, we are effectively performing tokenization twice. First, inks are converted to a model-agnostic sequence of indices in text format, as shown above. Then, these indices are converted to a final sequence of tokens accepted by the VLM by applying the model-specific tokenizer, e.g. using Byte-Pair Encoding. As a result, each token of the initial tokenization may correspond to several text tokens. Therefore, we observe that bigger indices coming out of the first tokenization result in increased sequence length of the second tokenization.

\subsection{Image representation}
\label{sec:method_image}

Rendering ink as an image enables conveying different information about the underlying digital ink, with the a simple approach being rendering black strokes on white background. Alternatively, \cite{KaggleQuickDraw} proposed to render time information in image channels. 

An added complexity in representing images of handwriting is the varying size and dimensions. There are several solutions to account for these, including stretching the sample to square, rendering respecting the aspect ratio, writing the image in multiple lines~\cite{fuyu-8b}, or using vision encoders that support arbitrary input image dimensions such as Pix2Struct~\cite{lee2023pix2struct} or NaViT~\cite{dehghani2023patch}.

In our work we use the vanilla ViT as the most popular and standard vision encoder solution. We encode speed information in input color channels, and render the input ink in several lines. We describe each of those below, provide examples of components of the rendering in Fig.~\ref{fig:rendering} and validate this choice in Section~\ref{sec:results_image}.

\textbf{Rendering the time and distance information.} 
We can use the three available color channels to encode the writing direction and stroke order. To do so, we normalize the time information within the ink between 0 and 1. The example of such representation is shown in the left part of Fig.~\ref{fig:rendering}, and the appropriate expression is given below

\begin{align}\label{eq:speed_rendering}
c^{R}_{i, j} = \frac{t_{i, j} - t_{0, 0}}{\max t_{m, n}} ~~
c^{G}_{i, j} = \frac{|d x_{i, j}|}{\max |d x_{m, n}|} ~~
c^{B}_{i, j} = \frac{|d y_{i, j}|}{\max |d y_{m, n}|}
\end{align}

where $t_{i,j}$ is the time value for the point $j$ in stroke $i$ and $d x_{i, j} = x_{i, j+1} - x_{i, j}$. Distance in $x$ and $y$ brings additional information about the size of the step and together with time information the model can estimate the speed. They are normalized between 0 and 1 similar to the time information. An example of this time and distance representation is shown in Fig.~\ref{fig:rendering}.

We show the importance of rendering both time and distance information in Section~\ref{sec:results_image} where we compare it to rendering time information only in 3 color channels or using simple black-on-white rendering.

\textbf{Rendering samples in multiple lines.} Since handwritten samples often have very skewed aspect ratio, rendering them in a single line on a fixed size image can often produce unreadable results. To account for this fact, they could be rendered on an image with aspect ratio $1:X$, which is then split horizontally and merged vertically to produce a single square image with multiple lines, see Fig.~\ref{fig:rendering}, far right. In our experiments, we use $X=2$ and validate this choice in Section~\ref{sec:results_image}.

\subsection{Target representation}
There are multiple ways to represent the label of handwriting, that needs to be predicted (the target). Having Byte-Pair Encoding tokenization in VLMs allows for faster decoding, but can be suboptimal for recognition performance, where two similar inputs (ex. ``hello" and ``hallo") may correspond to two different sequences of tokens. This is especially true in case of non-vocabulary words recognition. This can be circumvented by training the model to predict the target label broken into visual blocks of interest - ex. letters in case of text, of \LaTeX \  symbols in case of math recognition.

To achieve this, in case of text recognition we use space-separated letters as the target label (ex. ``h e l l o") to map visual elements (letters) to output tokens. For math expression recognition we don't use space separation as it allows a model to utilize knowledge about \LaTeX \ syntax from pretraining.  We validate this decision in Section~\ref{sec:target_representation}.

%% file: results.tex
\section{Experiments}
We evaluate the proposed representation of online handwriting on two different VLMs and compare results to a series of baselines. In our experiments we find that 
the model can fall back to the image representation when the text representation of handwriting is too long. Through ablation studies we find  the relative coordinate tokenizer, with text and speed rendering in images as the best-performing combination.

\subsection{Models}
\label{sec:results_models}

As described in section~\ref{sec:backdround}, we use two main base model families, PaLI~\cite{chen2023pali} combining ViT vision encoder with mT5 text encoder-decoder, and PaLM-E~\cite{driess2023palme}, projecting ViT tokens into a PaLM~\cite{anil2023palm} text decoder. For PaLI, we use ViT-B/16 vision encoder pretrained on JFT-300M~\cite{sun2017revisiting} and mT5-base pretrained on Common Crawl-based dataset, totalling 700M params.
For PaLM-E, we use ViT-B/16 vision encoder pretrained on CoCa~\cite{yu2022coca} and PaLM 128M text decoder, totalling 500M params, pretrained on a mix of social media, webpages, books and Github~\cite{anil2023palm}.

Additionally, results for LoRA-tuning reported in Table~\ref{table:lora_prompt} are based on a 5B parameter model of the same architecture as PaLI~\cite{chen2023pali3}. In spite of not being directly comparable to our smaller models, these results showcase the effectiveness of our method when applied through parameter-efficient tuning on larger models, where fine-tuning becomes computationally expensive. It is for this same reason that fine-tuning experiments of these larger model are omitted in this paper.

We compare our method to several baselines ranging from OCR models trained on private datasets to CTC transformer baseline which we train ourselves on the same public datasets.

\textbf{State-of-the-art methods.}  First, we compare to a publicly available OCR API~\cite{GoogleOcr}. For online handwriting recognition, we compare to~\cite{carbune2020recognition} which uses a 5M parameters LSTM encoder with a CTC decoder~\cite{10.1145/1143844.1143891}, combined with a character-based language model. Additionally, for the VNOnDB dataset, we report the results from 1) the best-performing method from VOHTR-2018 online handwriting recognition challenge~\cite{nguyen2018icfhr} which is Vietnamese-specific and 2) the OCR results of~\citet{le2019end}.

\textbf{Trained baselines.} We also compare the results to a Transformer encoder with a CTC decoder, similar to~\cite{alwajih2022transformer}. We provide the architecture details in Appendix ~\ref{sec:appendix_arch}.

\subsection{Datasets}
\begin{figure}[!ht]
    \centering
    \includegraphics[width=.99\linewidth]{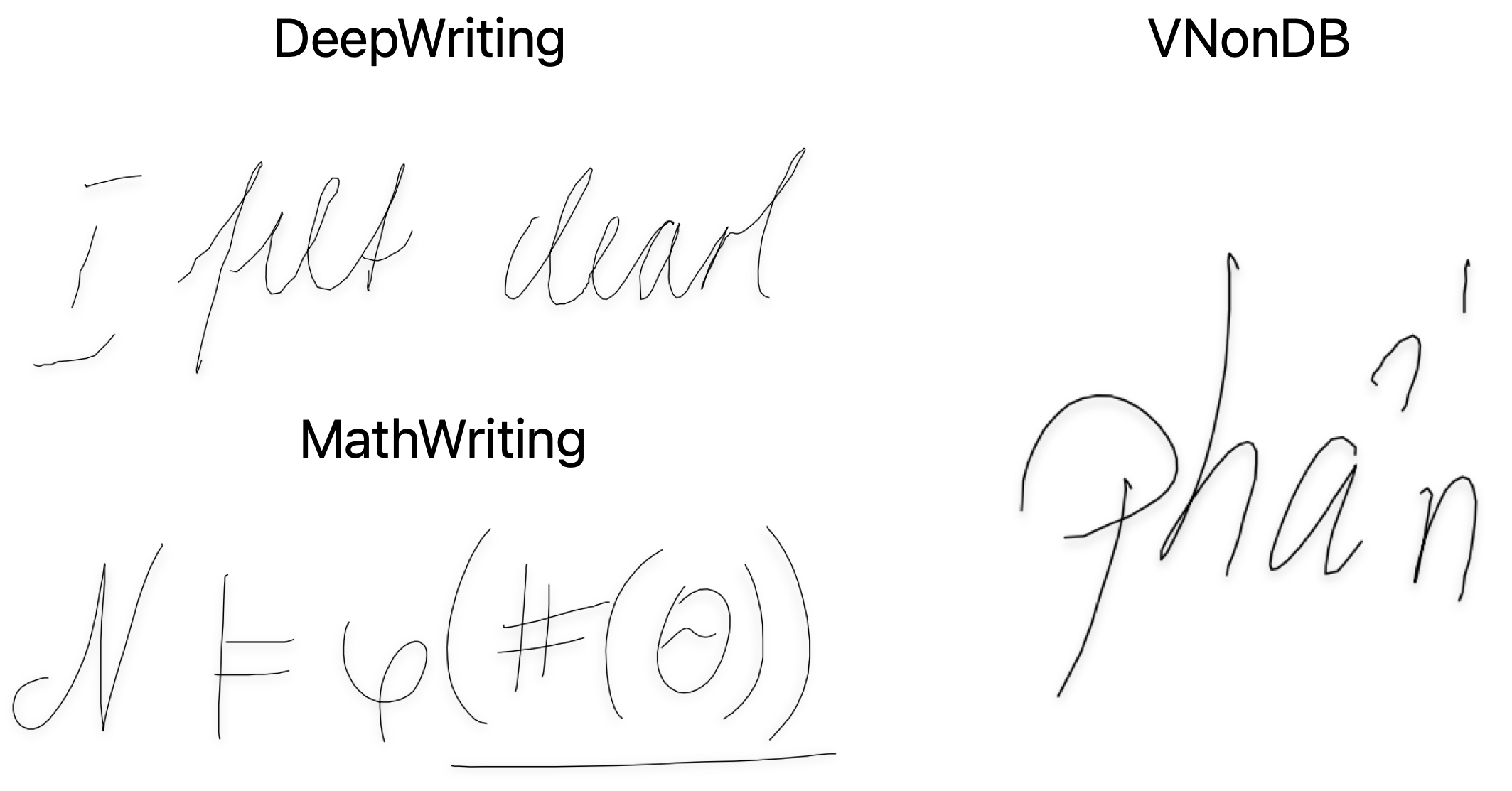}
    \caption{Examples from DeepWriting, MathWriting and VNonDB datasets.}
    \label{fig:dataset_examples}
\end{figure}

We train the models on the three public datasets – DeepWriting~\cite{aksan2018deepwriting}, MathWriting~\cite{MathWriting} and VNonDB~\cite{nguyen2018icfhr}. Dataset statistics are given in Table~\ref{tab:datasets} and examples from each dataset are shown in Fig.~\ref{fig:dataset_examples}.

\begin{table}[!ht]
    \caption{Dataset statistics. For MathWriting, 630k training samples includes 230k real and 400k synthetic samples.}
    \label{tab:datasets}
    \centering
    \footnotesize
    \begin{tabular}{l|c|c|c}
    Dataset & Language & Samples & Mean target tokens \\
    \hline
    DeepWriting & English & 34k & 13 \\
    MathWriting & \LaTeX & 630k & 15 \\
    VNOnDB & Vietnamese & 67k & 3 \\
    \end{tabular}
\end{table}

\subsection{Training and evaluation setup}

PaLI and PaLM-E models described in  Section~\ref{sec:results_models} are finetuned for 200k steps with 128 and 64 batch size respectively. CTC transformer is trained for 100k steps with 256 batch size. We use a context length 1024 for PaLI and 1500 for PaLM-E covering 95th percentile of sequence lengths in all validation datasets. These numbers differ between the two models due to different text tokenizers used in each, see details in Section~\ref{sec:backdround}. We use an additional 256 tokens coming from the ViT encoder on 224px image in case of PaLI and 288px in case of PaLM-E. During inference, we use greedy decoding with at most 64 tokens.

For PaLI and PaLM-E, we train on a 80\% MathWriting, 10\% VNOnDB, 10\% DeepWriting  mixture of datasets (a distribution similar to proportion of the number of samples in each dataset) as our representation allows seamless mixing of tasks and it is beneficial for the model quality. We additionally report numbers when trained on all datasets separately in the Appendix\ref{sec:mix_training}.

Similar to most literature on the topic, we report the standard Character Error Rate (CER) metric~\cite{michael2019evaluating}. We report the mean and variance over three runs for all the methods that we train. In the case of MathWriting we calculate CER based on the dictionary of \LaTeX \ tokens (see Appendix~\ref{sec:math_vocabulary}). In cases where we use space-separated target representation, we remove the spaces before CER computation.

\subsection{Results}
\label{sec:results}
\input{main_table}

Table~\ref{table:main_result} compares PaLI and PaLM-E models finetuned with our representation described in Section~\ref{sec:representation} (relative coordinates in text and speed rendering with two lines in image representation) to CTC transformer and state-of-the-art OCR and online handwriting recognition models. Our main conclusion from it is that {\bf performance of both VLMs finetuned for handwriting recognition is comparable to or better than state-of-the-art results.} Furthermore, best results for each dataset are achieved with VLMs, with the exception of VNOnDB. For VNOnDB, the best performing model is, unlike all the other approaches, tailored to Vietnamese -  it relies on specific handling of diacritics, Vietnamese-specific tokenizer, etc. We attribute the somewhat low performance of PaLM-E on Deepwriting to the small size of this dataset.

\input{lora_table}

{\bf Our proposed representation is useful in both fine-tuning and parameter-efficient tuning}. As visible in Table~\ref{table:lora_prompt}, fine-tuning with the 700M parameter model yields better results than LoRA-tuning 5B parameter model. However, LoRA-tuning with our representation performs better than image-based methods from Table~\ref{table:main_result}.

%% file: main_table.tex
\begin{table*}[!ht]
\caption{Comparison of our approaches to state-of-the-art methods on three public datasets. PaLI and PaLM-E are finetuned on relative coordinates in text and speed rendering in image representations. For VNonDB, DeepWriting we use space-separated target and for MathWriting targets are not space separated (see Sec.~\ref{sec:target_representation} for more details).}
\label{table:main_result}
\small
\centering
\begin{tabular}{ l|l|l|l|c|c|c }
& & & & \multicolumn{3}{c}{CER $\downarrow$} \\
 Model  & Training data & Language-specific & Input & MathWriting & VNOnDB & DeepWriting\\
  \hline
 \cite{GoogleOcr} & Private & - & Image & 5.93 & 4.83 & 14.16 \\
 \cite{le2019end} & Public  & No & Image &  - & 4.10 & - \\
  \hline
 \cite{carbune2020recognition} & Private & No & Ink &  - & 4.13 & 6.14 \\
 MyScript\cite{nguyen2018icfhr} & Private & Yes & Ink &  - & \textbf{2.91} & - \\
 CTC transformer & Public & No & Ink & 4.28 (0.06) & 3.82 (0.24) & 5.71 (0.2) \\
  \hline
 
 \textbf{PaLI [ours]} & \multirow{2}{*}{Public} & \multirow{2}{*}{No} & \multirow{2}{*}{Ink+Image}& 4.47 (0.07)  &  3.04 (0.01) & \textbf{4.39 (0.06)} \\
 \textbf{PaLM-E [ours]} & & & & \textbf{4.19 (0.04)} & 3.27 (0.1) & 6.89 (0.06) \\
\end{tabular}
\end{table*}

%% file: lora_table.tex
\begin{table}[!ht]
\caption{Comparison between finetuning the 700M PaLI and LoRA-tuning PaLI-5B with proposed ink+image representation.}
\label{table:lora_prompt}
\smallskip
\footnotesize
\centering
\begin{tabular}{p{7mm}|p{10mm}|c|c|c } 
 Train & \# Train. & \multicolumn{3}{c}{CER $\downarrow$} \\
 setup & params & MathWriting & VNOnDB & DeepWriting\\
  \hline
 Full & 700M & \textbf{4.47 (0.07)}  &  \textbf{3.04 (0.01)} & \textbf{4.39 (0.06)} \\
  LoRA & 27M & 4.93 (0.08)  & 3.61 (0.16) & 5.74 (0.17) \\
\end{tabular}
\end{table}

%% file: ablation.tex
\subsection{Ablation study}
In the following subsections we perform most of the ablation studies on the MathWriting dataset, which we chose due to its scale. We train PaLI for 100k steps in these experiments for shorter experiment cycles.

\subsubsection{Multimodal input}
\label{sec:multimodal}

\begin{table}[!ht]
\caption{Multimodality in PaLI and PaLM-E. In addition, we compare PaLI with 1024 and 512 tokens to see how image helps when ink is too long for context length. PaLM-E has a specific number tokenization, described in Section~\ref{sec:backdround} and requires a longer sequence length than PaLI. Ink and image representations are the same as in Table~\ref{table:main_result}.}
\label{table:multimodal}
\small
\centering
\begin{tabular}{ l|l|l|l|c } 
 &  & \multicolumn{2}{c|}{Seq. len} & CER $\downarrow$ \\
 Model  & Input & Ink & Image & MathWriting \\
 \hline
 \multirow{5}{*}{PaLI} & Image & - & 256 & 8.07 (0.14) \\
  & \multirow{2}{*}{Ink} & 1024 & - & 4.64 (0.07) \\
  &  & 512 & - & 10.65 (0.34) \\
  & \multirow{2}{*}{Ink + Image} & 1024 & 256 & \textbf{4.55 (0.04)} \\
  &  & 512 & 256 & 5.89 (0.35) \\
 \hline
  \multirow{3}{*}{PaLM-E}
  & Image & - &  256 & 4.87 (0.07) \\
  & Ink &  1500 & - & 6.46 (0.02) \\
   & Ink+Image & 1500 & 256 & \textbf{4.22 (0.16)} \\
\end{tabular}
\end{table}

\begin{figure*}[!ht]
    \centering
    \includegraphics[width=0.95\textwidth]{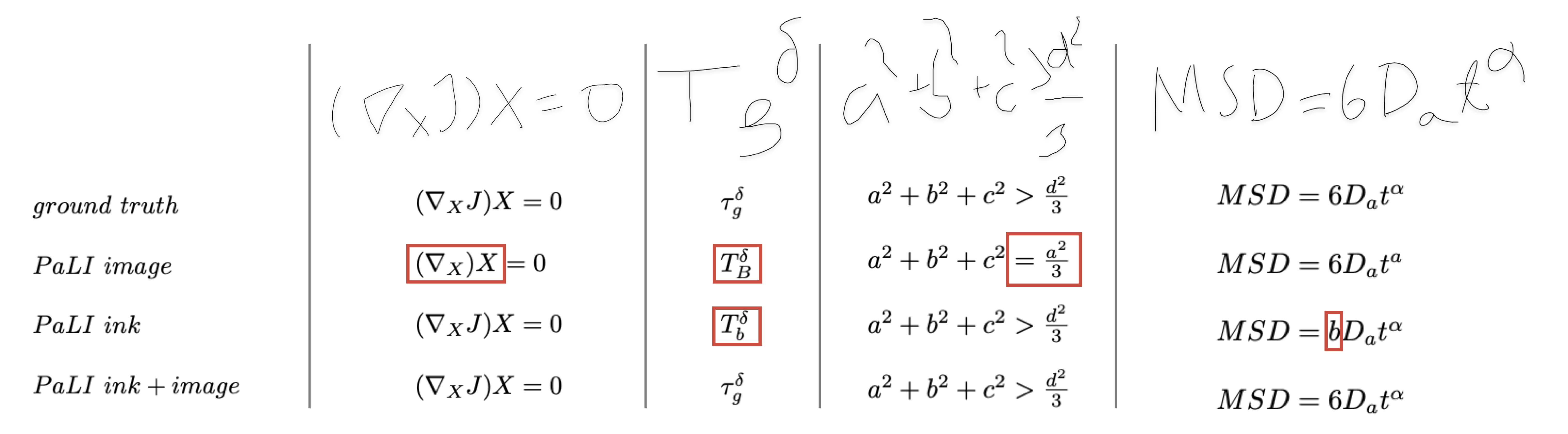}
    \vspace{-13pt}
    \caption{PaLI recognition on four examples where prediction only on image or ink is different from the target. We compare PaLI results to the ground truth from the MathWriting dataset. Mistakes include mixing similar characters like ``tau'' and ``T'', ``d'' and ``a''. We show that those mistakes are addressed by including ink representation. }
    \label{fig:reco_examples}
\end{figure*}

Experiments in Table~\ref{table:multimodal} show how sequence and image representations perform in recognition training with VLMs and when it is beneficial to use them in the representation. We show examples of mistakes in image-only recognition and show how they are addressed with an additional ink modality Fig.~\ref{fig:reco_examples}. We draw several conclusions:  (1) {\bf Combining both image and ink input yields better performance than when using only one of the input modalities}. This is particularly visible in the case of PaLM-E. Our hypothesis is that, due to the longer sequence length from the different tokenization of numbers in PaLM-E (see Section~\ref{sec:backdround}), training only on ink becomes a more challenging task than in PaLI. In this case addition of image tokens significantly improves the quality of the final model. (2) {\bf The input image helps if the sequence representation of the ink exceeds the context length}. We calculated that for the inks that exceed the sequence length CER goes down from 17.38 to 11.17 with an introduction of an image input for PaLI. 

\subsubsection{Ink tokenization}
\label{sec:ablation_tokenization}
In this ablation study we compare different approaches to tokenization from the Section~\ref{sec:sequence_representation}. We compare them on the PaLI model with only sequence representation in order to evaluate the effect of tokenization without an influence from an image input. We draw the following main conclusions based on Table~\ref{table:tokenization}: (1) {\bf Representing ink with text or separate tokens yields similar performance to extendeding the vocabulary as long as the whole ink sequence fits into the model context and the dictionary is small}. For the histogram-based tokenizer (similar to ~\cite{ribeiro2020sketchformer}, more details in the Appendix~\ref{sec:histogram_tokenizer}), which uses a large dictionary, each index corresponds to several tokens when represented as text, resulting in worse performance; (2) {\bf Absolute and relative representation of the ink result in a similar recognition quality}. This is also visible in Table~\ref{table:absolute_relative}. We have observed that this is dataset-specific, with English and Vietnamese recognition performing better with relative coordinates and Math recognition with absolute coordinates. We attribute this to the fact that relative representation is shift-invariant and is easier to learn, but for math expressions, which don't have a linear structure, absolute coordinates are more important. 

\begin{table}[!ht]
\caption{PaLI with different ink representations and no image input. Models with extended dictionaries are compared to text representation of inks.}
\label{table:tokenization}
\small
\centering
\begin{tabular}{l|l|c } 
 & & CER $\downarrow$ \\
 Tokenization & Token type & MathWriting \\
  \hline
  \multirow{2}{*}{Relative} & Text & 4.64 (0.07) \\
  & Extended & 4.59 (0.08) \\
  \multirow{2}{*}{Absolute} & Text & \textbf{4.20 (0.02)} \\
   & Extended & 4.43 (0.04) \\
  \multirow{2}{*}{\cite{ribeiro2020sketchformer}} & Text & 7.11 (0.08) \\
   & Extended & 4.53 (0.04) \\
\end{tabular}
\end{table}

\begin{table}[!ht]
\caption{Absolute coordinates vs relative offsets for text representation combined with an image in the PaLI model. Absolute representation performs 12\% worse on DeepWriting and 4\% on VNOnDB.}
\label{table:absolute_relative}
\small
\centering
\begin{tabular}{p{15mm}|p{15mm}|p{15mm}|p{15mm} } 
 & \multicolumn{3}{c}{CER $\downarrow$} \\
 Tokenization & MathWriting & VNOnDB & DeepWriting\\
  \hline
 Relative  & 4.55 (0.04) &  \textbf{4.55 (0.09)} & \textbf{4.68 (0.16)} \\
 Absolute  & \textbf{4.37 (0.04)} & 4.72 (0.0) & 5.23 (0.1)\\
\end{tabular}
\end{table}

\subsubsection{Image rendering}
\label{sec:results_image}

\begin{table}[!ht]
\caption{Different rendering options w.r.t. rendering with color and in multiple lines for PaLI model. Measured with only image input to estimate the effect of image representation without an influence of ink representation. Examples of all options that we evaluate are shown in Fig.~\ref{fig:rendering}.
}
\label{table:speed_lines}
\small
\centering
\begin{tabular}{l|l|l|c } 
 \multicolumn{2}{c}{Color} & & CER $\downarrow$ \\
 Time & Distance & \# Lines & MathWriting\\
  \hline
 No & No & 2 & 13.93 (4.34) \\
 Yes & No & 2 & 11.36 (1.49)  \\
 Yes & Yes  & 2 & \textbf{8.07 (0.14)} \\
 \hline
  Yes & Yes & 1 & 11.15 (1.69) \\
  Yes & Yes  & 2 & \textbf{8.07 (0.14)} \\
  Yes & Yes & 4 & 14.94 (1.81) \\
\end{tabular}
\end{table}

In the following ablation study we compare three different types of color rendering – black\&white, time and time+distance; see examples in Fig.~\ref{fig:rendering}. We focus our attention specifically on time information as it is a distinguishing feature of online handwriting recognition task. From Table~\ref{table:speed_lines} we draw the conclusion that \textbf{time and distance information in color channels both contribute to a better performance}. In Appendix~\ref{sec:appendix_time_image} we present an example of how writing order helps with recognition of an ambiguous ink.

We then explore in Table~\ref{table:speed_lines} the optimal number of lines for rendering. Rendering in multiple lines allows to control the size of writing, which can be small if an ink is rendered in one line (see Fig.~\ref{fig:rendering}) or in too many lines.
 We find that two lines is optimal for MathWriting dataset which has a median aspect ratio of 2.29. We conclude that the \textbf{number of lines closest to average aspect ratio in the dataset performs the best}.

%%%%%%%%%%%%%%%%%%%%%%%%%%%%%%%%%%%%%%%%%%%%%%%%%%%%%%%%%%%%%%%%%%%%%%%%%%%%%%
%%%%%%%%%%%%%%%%%%%%%%%%%%%%%%%%%%%%%%%%%%%%%%%%%%%%%%%%%%%%%%%%%%%%%%%%%%%%%%
\subsubsection{Target representation}
\label{sec:target_representation}
In Table~\ref{table:spaces} we show that \textbf{space-separation of the target improves handwriting text recognition but performs poorly on mathematical expression recognition}. The gap in performance on DeepWriting is especially big as this dataset contains many non-vocabulary words like ``clearl'' in Fig.~\ref{fig:dataset_examples}. In case of MathWriting we obtain much better results without space interleaving as it allows the model to utilize knowledge about \LaTeX \ syntax from pretraining.

\begin{table}[!ht]
\caption{Comparison on space interleaving in target on PaLI relative ink+image. This target representation performs well on VNOnDB and DeepWriting datasets but decreases the quality on MathWriting where multiple characters in the target frequently represents one symbol in ink.}
\label{table:spaces}
\small
\centering
\begin{tabular}{p{7mm}|p{16mm}|p{15mm}|p{15mm} } 
 & \multicolumn{3}{c}{CER $\downarrow$} \\
Spaces & MathWriting & VNOnDB & DeepWriting\\
  \hline
 Yes  & 11.03 (0.36) & \textbf{4.55 (0.09)} & \textbf{4.68 (0.16)} \\
  No & \textbf{4.55 (0.04)} & 5.02 (0.26) & 22.3 (1.79) \\
\end{tabular}
\end{table}

%% file: related_work.tex
\section{Related work}
\label{sec:related_work}
The \textbf{online handwriting recognition} has a long history, and ink representation plays an important role in any recognition model. In early works inks were represented as aggregations of geometric features like direction, distances, curvatures, etc described in \citep{Japanese_online}. With more complex machine learning systems, manual feature engineering became unnecessary as models with sufficient data can learn relevant features from raw representations, for instance in computer vision \citep{NIPS2012_c399862d} and in NLP \citep{NIPS2013_9aa42b31}.

One of the main challenges in online handwriting recognition is aligning the input ink with the target text as they have different lengths. Historically, Hidden Markov Models were used for this task \cite{541414}. They were subsequently replaced by deep neural networks like LSTMs \cite{10.1162/neco.1997.9.8.1735} and Transformers \cite{DBLP:journals/corr/VaswaniSPUJGKP17} with \textbf{connectionist temporal classification} (CTC) loss. This solution was initially proposed for speech recognition in \cite{10.1145/1143844.1143891} and it trains the segmentation and classification together. Another approach is to use the encoder-decoder model (also known as seq2seq) where an encoder produces a sequence of states based on the input and an autoregressive decoder uses those states in cross-attention. This method has been widely adopted in speech recognition in recent years \cite{Prabhavalkar2017ACO,Pundak2018DeepCE, Chiu2017StateoftheArtSR, Hannun2014DeepSS}.
Similarly, decoder-only models like PaLM \cite{anil2023palm} can be adopted to speech recognition \cite{rubenstein2023audiopalm}.
In \textbf{offline handwriting recognition} the input is a scanned image or a photo of handwriting and the output is digital text. Modern approaches to offline handwriting recognition are mainly based on deep neural nets according to a survey on handwritten OCR \cite{survey_handwriting}.

The idea to \textbf{combine online and offline recognition} has been explored before in \cite{8978103}, which proposed a special variation of encoder-decoder architecture to combine sequence data and image representation. A similar idea for sketch generation was investigated in \cite{pourreza2023painter} where the image representation was merged with an LLM using a cross-attention mechanism.

In recent years \textbf{Large Language Models} have demonstrated impressive capabilities in various tasks like question answering, math problem solving, summarization etc. \cite{anil2023palm, brown2020language, xue2021mt5}. Recognizing the limitations of text-only models, researchers are now exploring how integrating different modalities like vision can unlock even greater capabilities for large language models. The main goal of this expansion is to have a universal model \cite{bommasani2022opportunities} that can be easily adapted to a variety of tasks like image captioning \cite{zhou2019unified}, speech translation \cite{huang2023speech} and many more. The models that combine visual and language components (\textbf{Large Visual Language Models}) include Flamingo \cite{alayrac2022flamingo}, PaLI \cite{chen2023pali, chen2023palix}, PaLM-E \cite{driess2023palme}, LLaVA \cite{liu2023visual}. Those models vary in size and training datasets, but most notably in architectural approaches to combining image and text modalities. In Flamingo gated cross-attention was used to connect language and image modalities. PaLI expands the encoder state with image embeddings and utilizes cross-attention in an encoder-decoder to process images and text together. PaLM-E and LLaVA models rely on self-attention as they fuse projected image embeddings into the sequence of text tokens. 

\textbf{Our work} We argue that VLMs provide a natural framework to combine online and offline representations of digital inks. Unlike \cite{8978103} we utilize already existing VLMs for easier processing of both ink sequences and images. Our method is focused on finding text and image representations that work best with VLMs.

%% file: conclusion.tex
\section{Conclusion}
In this work, we provide a representation of handwriting for VLMs as a text sequence and an image that enables tuning VLMs to a quality comparable to state of the art on three public datasets. Our representation includes relative discretized coordinates as text and rendered ink with time and distance information as image.

We show that (1) VLMs profit from multimodal inputs and that the image is particularly important in cases where the ink's text representation doesn't fit into context length (2) multiple handwriting tasks can be combined with this representation and (3) it is compatible with parameter-efficient tuning as well as fine-tuning. This points us to a future direction of exploring different handwriting task combinations in large VLMs.

\section{Impact Statement}
This paper presents work whose goal is to advance the field of Machine Learning. There are many potential societal consequences of our work, none which we feel must be specifically highlighted here.

\section{Acknowledgements}
We would like to thank Philippe Gervais, Sean Kirmani, Henry Rowley, Blagoj Mitrevski, Arina Rak, Julian Schnitzler, Chengkun Li, Vincent Etter for their helpful suggestions.

%% file: appendix.tex
\section{Time sampling}
\label{sec:appendix_time_sampling}
\begin{figure}[!ht]
    \centering
    \includegraphics[width=\textwidth/2]{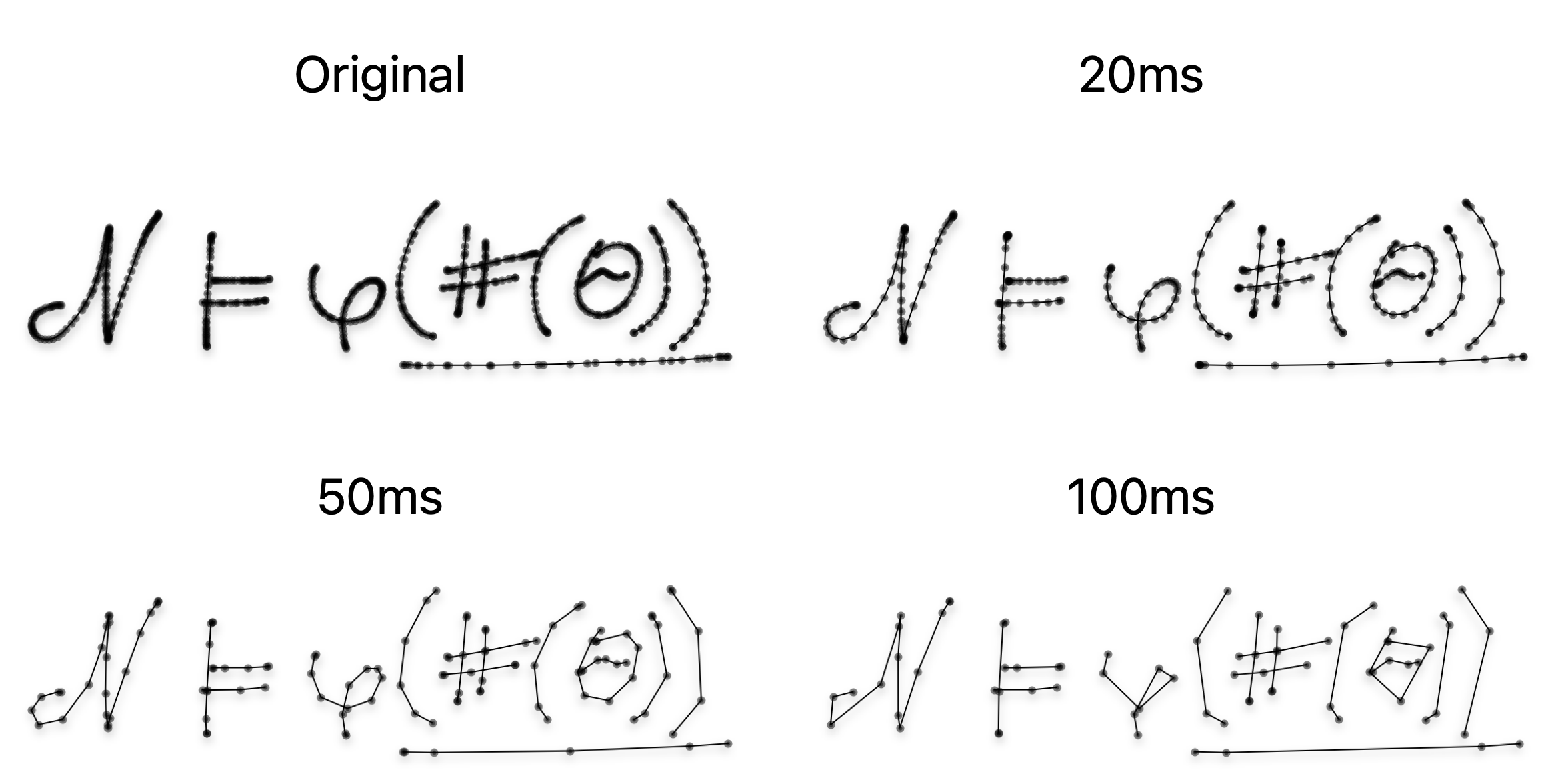}
    \caption{Different time sampling deltas. In our experiments we use 20ms for all datasets. Importantly, note that larger deltas result in a shorter representation.}
    \label{fig:time_sampling}
\end{figure}
In Fig.~\ref{fig:time_sampling} we show how the time sampling delta affects the ink. With bigger deltas we get less points in the ink and important details of writing can be missed.

\section{Training on the mix of datasets}
\label{sec:mix_training}

\begin{table}[!ht]
\caption{A comparison of fine-tuning separately on each dataset or on the mix of 80\% MathWriting, 10\% DeepWriting and 10\% VNOnDB.}
\label{table:combined}
\footnotesize
\centering
\begin{tabular}{ l|l|c|c|c } 
  & & \multicolumn{3}{c}{CER $\downarrow$} \\
 model & mix data & MathWriting & VNOnDB & DeepWriting\\
  \hline
  CTC transformer & \multirow{3}{*}{no} & 4.28 (0.06) & 3.82 (0.24) & 5.71 (0.2)\\
 PaLI &   & 4.36 (0.04) &  4.55 (0.09) & 4.68 (0.16) \\
 PaLM-E &  &  4.22 (0.16) & 3.42 (0.05) & 7.98 (0.17) \\
 \hline
 CTC transformer & \multirow{3}{*}{yes} & 5.31 (0.03) & 4.33 (0.04) & 6.12 (0.24) \\
 PaLI &  & 4.47 (0.07)  &  \textbf{3.04 (0.01)} & \textbf{4.39 (0.06)} \\
 PaLM-E & & \textbf{4.19 (0.04)} & 3.27 (0.1) & 6.89 (0.06) \\
\end{tabular}
\end{table}
In Table~\ref{table:combined} we show that both PaLI and PaLM-E models benefit from training on a mix of datasets whereas CTC transformer doesn't. VNOnDB and DeepWriting character error rate decreases significantly if a mix of datasets is used with VLMs. We observe the decrease of PaLI MathWriting quality when trained on the mix of datasets which we attribute to fewer math tokens observed during training. We also show that training CTC transformer model on a mix of datasets doesn't lead to improvements on any of the datasets. Due to this fact we report CTC transformer that was trained on separate datasets in the Section~\ref{sec:results}.  

\section{Time and distance rendering in images}
\label{sec:appendix_time_image}
\begin{figure}[!ht]
    \centering
    \includegraphics[width=\textwidth/3]{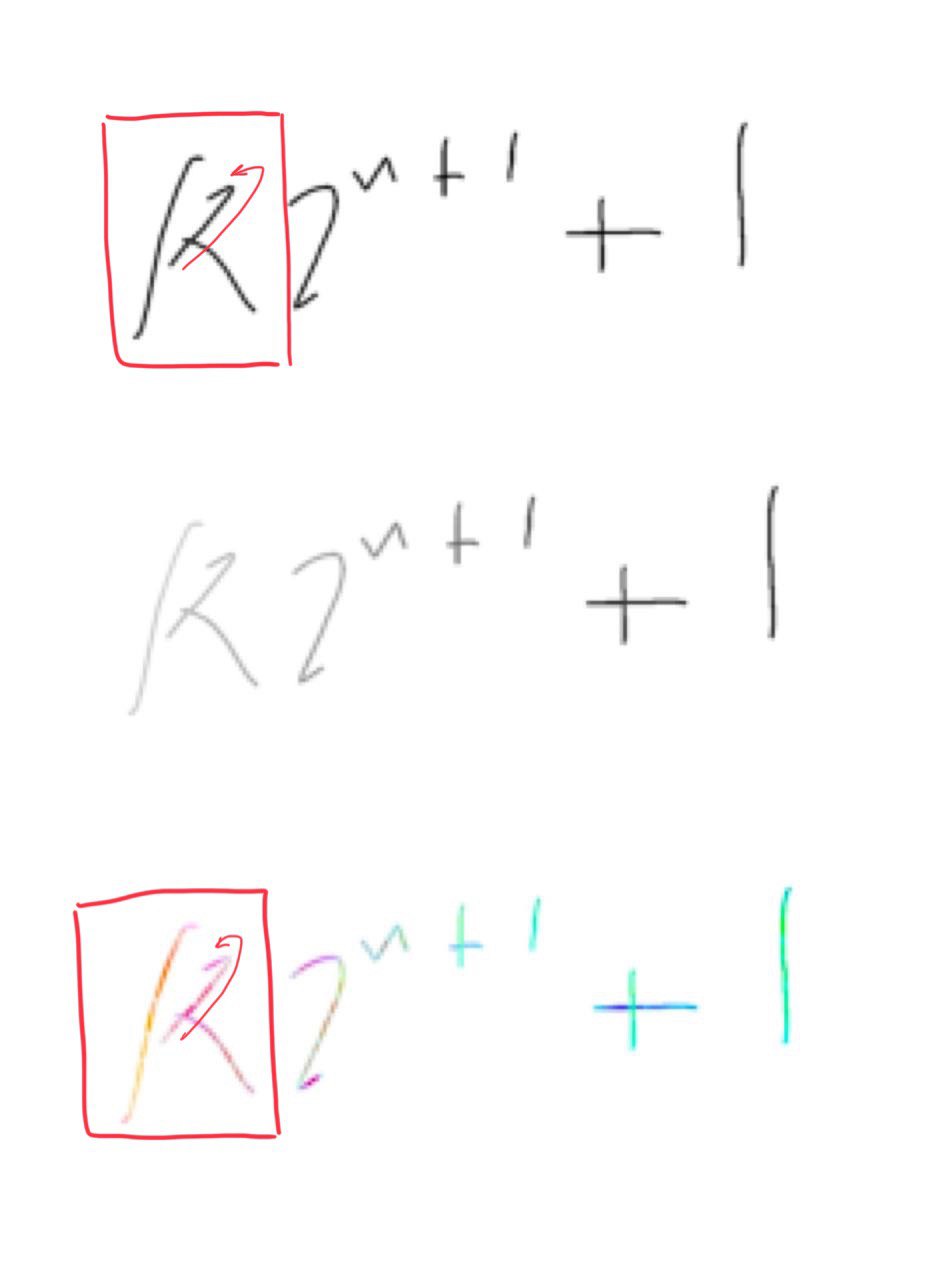}
    \caption{Example with an ambiguous first letter. Depending on the order of points it might be ``K'' or ``R''. With time and distance information in the image our model changes the wrong prediction ``R'' to correct ``K''. }
    \label{fig:hard_example_annotation}
\end{figure}
In Fig.~\ref{fig:hard_example_annotation} we provide an example of ambiguous writing where order of writing helps a model to recognize the ink correctly.

\section{Baseline architecture}
\label{sec:appendix_arch}

\begin{table}[!ht]
\caption{CTC transformer architecture details.}
\label{table:ctc_transformer_detail}
\footnotesize
\centering
\begin{tabular}{ l|l|l|l } 
  & MathWriting & VNOnDB & DeepWriting\\
  \hline
  \textbf{parameters} & 35M & 15M & 9M \\
  layers & 11 & 7 & 10\\
  embedding size & 512 & 384 & 192 \\
  attention heads & 8 & 8 & 2\\
  units per head & 256 & 256 & 1024\\
  activation & swish & gelu & gelu\\
  dropout & 0.15 & 0.1 & 0.2 \\
\end{tabular}
\end{table}

Our baseline architecture is a Transformer encoder combined with a CTC loss similar to \cite{alwajih2022transformer}. We provide the details of models for each dataset in Table.~\ref{table:ctc_transformer_detail}. The raw points from the ink are encoded into Bézier curves to account for factors such as input device sample rate and length of the resulting input features \cite{carbune2020recognition}, therefore removing the need for further preprocessing of the ink. The input features are then processed by a transformer encoder with multiple attention layers and a final logit layer of the size of the vocabulary. We train the resulting model end-to-end using a CTC loss. We don't combine a trained recognizer model with an additional external language model (ex. in \cite{carbune2020recognition}) on top of transformer due to lack of public training data for this process.

\section{Histogram tokenizer}
\label{sec:histogram_tokenizer}

\begin{table}
\caption{Coordinate vs histogram tokenizer.}
\label{table:coordinate_histogram}
\small
\centering
\begin{tabular}{ |l|c|c| } 
 \hline
   & coordinate & histogram \\
 \hline
  resize & yes & no \\
 \hline
  vocabulary & 900 & 12k \\
 \hline
  tokens per point & 2 & 1 \\
 \hline
\end{tabular}
\end{table}

We compare our method shown in Fig.~\ref{fig:coordinate_tokenization} to a method where one point translates into one token in the sequence see Table~\ref{table:coordinate_histogram}. That is especially important to consider because bigger sequence length requires more resources and can lead to lower performance in LLMs \cite{anil2022exploring}. We can potentially enumerate all bins in coordinate tokenizer which would result in $N^2$ vocabulary size. However, for size $N$ big enough to have good reconstruction (224 in our experiments) we will end up with at least 50k tokens. For this reason, we train a histogram tokenization method based on the polar form of offsets – log distance and angle. In this approach we don't use scale normalization to a fixed-size canvas as it is not required by the method. During training we split angles uniformly across $2\pi$ into $100$ buckets and than we iteratively do a binary split of log distance intervals until they contain less than $0.1\%$ of train data. This way we end up with much smaller buckets near $0$ as they are more frequent.

\section{MathWriting vocabulary}
\label{sec:math_vocabulary}
We use a special vocabulary for MathWriting dataset to account for the fact that some math symbols are represented in text with multiple characters, for instance $\theta$ would appear in target as ``theta''. This vocabulary consists of 142 special latex symbols and 87 English characters with punctuation.

``!'', ``\&'', ``('', ``)'', ``*'', ``+'', ``,'', ``-'', ``.'', ``/'', ``0'', ``1'', ``2'', ``3'', ``4'', ``5'', ``6'', ``7'', ``8'', ``9'', ``:'', ``;'', ``<'', ``='', ``>'', ``?'', ``A'', ``B'', ``C'', ``D'', ``E'', ``F'', ``G'', ``H'', ``I'', ``J'', ``K'', ``L'', ``M'', ``N'', ``O'', ``P'', ``Q'', ``R'', ``S'', ``T'', ``U'', ``V'', ``W'', ``X'', ``Y'', ``Z'', ``['', ``\textbackslash \#'', ``\textbackslash \%'', ``\textbackslash \&'', ``\textbackslash Delta'', ``\textbackslash Gamma'', ``\textbackslash Lambda'', ``\textbackslash Leftrightarrow'', ``\textbackslash Omega'', ``\textbackslash Phi'', ``\textbackslash Pi'', ``\textbackslash Psi'', ``\textbackslash Rightarrow'', ``\textbackslash Sigma'', ``\textbackslash Theta'', ``\textbackslash Upsilon'', ``\textbackslash Vdash'', ``\textbackslash Xi'',
``\textbackslash '', ``\textbackslash \_'', ``\textbackslash aleph'', ``\textbackslash alpha'', ``\textbackslash angle'', ``\textbackslash approx'', ``\textbackslash backslash'', ``\textbackslash begin{matrix}'', ``\textbackslash beta'', ``\textbackslash bigcap'', ``\textbackslash bigcirc'', ``\textbackslash bigcup'', ``\textbackslash bigoplus'', ``\textbackslash bigvee'', ``\textbackslash bigwedge'', ``\textbackslash bullet'', ``\textbackslash cap'', ``\textbackslash cdot'', ``\textbackslash chi'', ``\textbackslash circ'', ``\textbackslash cong'', ``\textbackslash cup'', ``\textbackslash dagger'', ``\textbackslash delta'', ``\textbackslash div'', ``\textbackslash dot'', ``\textbackslash emptyset'', ``\textbackslash end{matrix}'', ``\textbackslash epsilon'', ``\textbackslash equiv'', ``\textbackslash eta'', ``\textbackslash exists'', ``\textbackslash forall'', ``\textbackslash frac'', ``\textbackslash gamma'', ``\textbackslash ge'', ``\textbackslash gg'', ``\textbackslash hat'', ``\textbackslash hbar'', ``\textbackslash hookrightarrow'', ``\textbackslash iff'', ``\textbackslash iint'', ``\textbackslash in'', ``\textbackslash infty'', ``\textbackslash int'', ``\textbackslash iota'', ``\textbackslash kappa'', ``\textbackslash lambda'', ``\textbackslash langle'', ``\textbackslash lceil'', ``\textbackslash le'', ``\textbackslash leftarrow'', ``\textbackslash leftrightarrow'', ``\textbackslash lfloor'', ``\textbackslash ll'', ``\textbackslash longrightarrow'', ``\textbackslash mapsto'', ``\textbackslash mathbb'', ``\textbackslash models'', ``\textbackslash mp'', ``\textbackslash mu'', ``\textbackslash nabla'', ``\textbackslash ne'', ``\textbackslash neg'', ``\textbackslash ni'', ``\textbackslash not'', ``\textbackslash notin'', ``\textbackslash nu'', ``\textbackslash odot'', ``\textbackslash oint'', ``\textbackslash omega'', ``\textbackslash ominus'', ``\textbackslash oplus'', ``\textbackslash otimes'', ``\textbackslash overline'', ``\textbackslash partial'', ``\textbackslash perp'', ``\textbackslash phi'', ``\textbackslash pi'', ``\textbackslash pm'', ``\textbackslash prime'', ``\textbackslash prod'', ``\textbackslash propto'', ``\textbackslash psi'', ``\textbackslash rangle'', ``\textbackslash rceil'', ``\textbackslash rfloor'', ``\textbackslash rho'', ``\textbackslash rightarrow'', ``\textbackslash rightleftharpoons'', ``\textbackslash sigma'', ``\textbackslash sim'', ``\textbackslash simeq'', ``\textbackslash sqrt'', ``\textbackslash sqsubseteq'', ``\textbackslash subset'', ``\textbackslash subseteq'', ``\textbackslash subsetneq'', ``\textbackslash sum'', ``\textbackslash supset'', ``\textbackslash supseteq'', ``\textbackslash tau'', ``\textbackslash theta'', ``\textbackslash tilde'', ``\textbackslash times'', ``\textbackslash top'', ``\textbackslash triangle'', ``\textbackslash triangleleft'', ``\textbackslash triangleq'', ``\textbackslash underline'', ``\textbackslash upsilon'', ``\textbackslash varphi'', ``\textbackslash varpi'', ``\textbackslash varsigma'', ``\textbackslash vartheta'', ``\textbackslash vdash'', ``\textbackslash vdots'', ``\textbackslash vec'', ``\textbackslash vee'', ``\textbackslash wedge'', ``\textbackslash xi'', ``\textbackslash zeta'', ``\textbackslash $\{$'', ``\textbackslash |'', ``\textbackslash $\}$'', ``]'', ``\textasciicircum'', ``\_'', ``a'', ``b'', ``c'', ``d'', ``e'', ``f'', ``g'', ``h'', ``i'', ``j'', ``k'', ``l'', ``m'', ``n'', ``o'', ``p'', ``q'', ``r'', ``s'', ``t'', ``u'', ``v'', ``w'', ``x'', ``y'', ``z'', ``$\{$'', ``$\vert$'', ``$\}$''